\documentclass[sigconf]{acmart}

\usepackage{amsmath}
\usepackage{algorithm}
\usepackage{algpseudocode}
\usepackage{array}
\usepackage{float} 
\usepackage{color}
\makeatletter
\def\BState{\State\hskip-\ALG@thistlm}
\makeatother

\begin{document}

\copyrightyear{2018} 
\acmYear{2018} 
\setcopyright{acmcopyright}
\acmConference[AISec '18]{11th ACM Workshop on Artificial Intelligence and Security}{October 19, 2018}{Toronto, ON, Canada}
\acmBooktitle{AISec '18: 11th ACM Workshop on Artificial Intelligence and Security Oct. 19, 2018, Toronto, ON, Canada}
\acmPrice{15.00}
\acmDOI{10.1145/3270101.3270111}
\acmISBN{978-1-4503-6004-3/18/10}

    \title{Dummy}

    \author{Mohammad Hashemi}
    \affiliation{%
        \institution{University of Colorado Boulder}
        \city{Boulder}
        \state{Colorado} 
    }
    \email{mohammad.hashemi@colorado.edu}

	\author{Greg Cusack}
    \affiliation{%
        \institution{University of Colorado Boulder}
        \city{Boulder}
        \state{Colorado} 
    }
    \email{gregory.cusack@colorado.edu}

    \author{Eric Keller}
    \affiliation{%
        \institution{University of Colorado Boulder}
		\city{Boulder}
        \state{Colorado} 
    }
    \email{eric.keller@colorado.edu	}

\newcommand{\blue}[1]{\textcolor{black}{#1}}
\newcommand{\red}[1]{\textcolor{black}{#1}}
\newcommand{\gr}[1]{\textcolor{black}{#1}}


\title{Stochastic Substitute Training: A Gray-box Approach to Craft Adversarial Examples Against Gradient Obfuscation Defenses} 

\begin{abstract}

It has been shown that adversaries can craft example inputs to neural networks which are similar to 
legitimate inputs but have been created to purposely cause the neural network to misclassify the input.
These adversarial examples are crafted, for example, by calculating gradients of a carefully defined loss function with respect to the input.  
As a countermeasure, some researchers have tried to design robust models by blocking or obfuscating gradients, even in white-box settings. 
Another line of research proposes introducing a separate detector to attempt to detect adversarial \gr{examples}.  \gr{This approach also makes} use of gradient obfuscation techniques, for example\gr{,} to prevent the adversary from trying to fool the detector.
In this paper, we introduce stochastic substitute training, a gray-box approach 
that can craft adversarial examples for defenses which obfuscate gradients.
For those defenses that \blue{have tried to make models more robust,}
with our technique, an adversary can craft adversarial examples with no knowledge of the defense.  
For defenses that attempt to detect the adversarial examples, 
with our technique, an adversary only needs very limited information about the defense to craft adversarial examples.
We demonstrate our technique by applying it against two defenses which make models more robust and two defenses which detect adversarial examples.
\end{abstract}




\maketitle
\section{Introduction}

Deep learning has evolved 
in many areas.
These deep neural networks show promising results in tasks such as malware detection \cite{droid}, autonomous driving \cite{drive}, network intrusion detection \cite{intrusion}, diagnosis in medical images\cite{tumor},
and in applications such as image classification, deep neural networks can even surpass human level performance. \cite{delv}
Deep reinforcement learning has also demonstrated promising results in recent years in many decision making problems, such as human level control in Atari video games \cite{atari}, defeating \gr{the} best human players in the game of Go \cite{Go}, making a humanoid robot run \cite{robot_control}, and resource management in a cluster with different resource types \cite{resource}.

Despite their success in a wide range of applications, deep neural networks, like other traditional classifiers, suffer from a vulnerability to adversarial examples.
When working with images, an adversarial example is an image which is carefully modified to make a classifier predict it incorrectly with minimal modifications to the image.
\gr{In many cases,} the perturbation that is added to these images \gr{is} imperceptible  to a human observer.  Therefore, a human is likely to classify the images as they did before the alterations.
\blue{This problem \gr{is} not limited to modifications to digital images. Kurakin et al. in \cite{kurakin} showed for the first time that this attack is applicable in the physical world as well. Later,} Eykholt et al. in \cite{advsign} showed that an adversary can place a few stickers on a stop sign to fool a classifier, \textit{e.g.,} causing it to predict the sign as a speed limit sign. 

Due to the threat that adversarial examples pose, many researchers have proposed solutions to address this vulnerability. These works fall into two main categorizations. In one line of work, \gr{researchers} introduced different mechanisms to make classifiers more robust to adversarial examples such that the models classify adversarial inputs that are visually close to legitimate inputs correctly ~\cite{therm, rfn}. We refer to these defenses as ``fortifying defenses''. In the other line of work, \gr{others} have tried to distinguish between legitimate examples and adversarial examples using some detection mechanisms ~\cite{defensegan, safetynet}. We refer to these defenses as ``detecting defenses''.  
\blue{One method an attacker can use to craft an adversarial example is to calculate the gradients of a loss function with respect to the input.} Carlini et al. in \cite{carlini_detection} showed that an adversary can bypass ten detection methods by changing this loss function. Since that time, both the detecting defenses and fortifying defenses have evolved. The defenses now leverage techniques that prevent the adversary from getting a useful gradient from the model or the detector even when the loss function is changed. \red{These techniques are called gradient masking as introduced by Papernot et al. in \cite{practical_bb}}.  Athalye et al. in \cite{obfuscated}, however, demonstrated these defenses are still vulnerable by showing a white-box approach to craft adversarial examples against these defenses, where the attacker needs to know about the defenses, their parameters, and model parameters.

In this paper, we introduce Stochastic Substitute training (SST), which is 
\blue{ an easy and general gray-box attack}\gr{, for breaking}  defenses \gr{that} obfuscate gradients without any knowledge about the model's parameters, the defense parameters, or access to the training dataset.  \blue{SST only assumes access to the logits (inputs of softmax layer) and doesn't need to be tailored \gr{to} different defenses in the case where they fortify a model. That is, for fortifying defenses\gr{,} SST is completely generic. For detecting defenses, the attacker should bring the detection conditions into the loop of crafting adversarial examples.}
We do this by training a substitute model with stochastically modified inputs. These inputs are the set of images that an adversary wants to craft adversarial examples for. 
We evaluate two fortifying defenses and two detecting defenses. But our approach is not limited to these defenses and can be applied to others as well. 

We make the following contributions:
\begin{itemize}
\item We introduce Stochastic Substitute Training (SST) to craft adversarial examples for models that obfuscate gradients, as old methods, such as those introduced in \citep{intriguing,fgsm,cw,jsma}, are ineffective in crafting adversarial examples for models that obfuscate gradients.
  \item We evaluate two fortifying defenses, random feature nullification (RFN) \cite{rfn} and thermometer encoding \cite{therm}, on the MNIST~\cite{mnist} and CIFAR-10~\cite{cifar} datasets respectively. With SST, we show that an adversary can craft adversarial examples for models fortified with these defenses with no knowledge about the defense and with a small amount of perturbation.
  \item We evaluate two detecting defenses, SafetyNet \cite{safetynet} and Defense-GAN \cite{defensegan}, on the MNIST dataset.  We show how an adversary can bypass these detection methods.
	\item \blue{We compare against two black-box attacks \citep{practical_bb,transferable} that can be used to evaluate defenses which obfuscate gradients without knowledge about the defense. 
We show that evaluating with these black-box approaches provides the defenses have a false sense of security. \gr{Since,} with the minimal extra visibility (the logits) in our gray-box approach,  we are 
capable of crafting adversarial examples that the black-box attacks cannot.}
\end{itemize}

The rest of this paper is organized as follows: in Section 2, we provide the reader with background information. Then, in Section 3, we introduce SST, our new approach for crafting adversarial examples for defenses which obfuscate gradients. In Section 4, we evaluate our approach against fortifying defenses, and in section 5, we evaluate our approach against detecting defenses. In Section 6, we compare against \blue{two black-box} attacks against the aforementioned defenses, and finally in Section 7, we conclude the paper.




\section{Background}

In this section\gr{,} we first briefly explain how deep neural networks work for image classification and then introduce the notation we use in this paper. \blue{Finally, we go over how an adversary can craft an adversarial example}. 

\subsection{Deep Neural Networks}
\begin{figure}
  \centering
  \includegraphics[width=80mm]{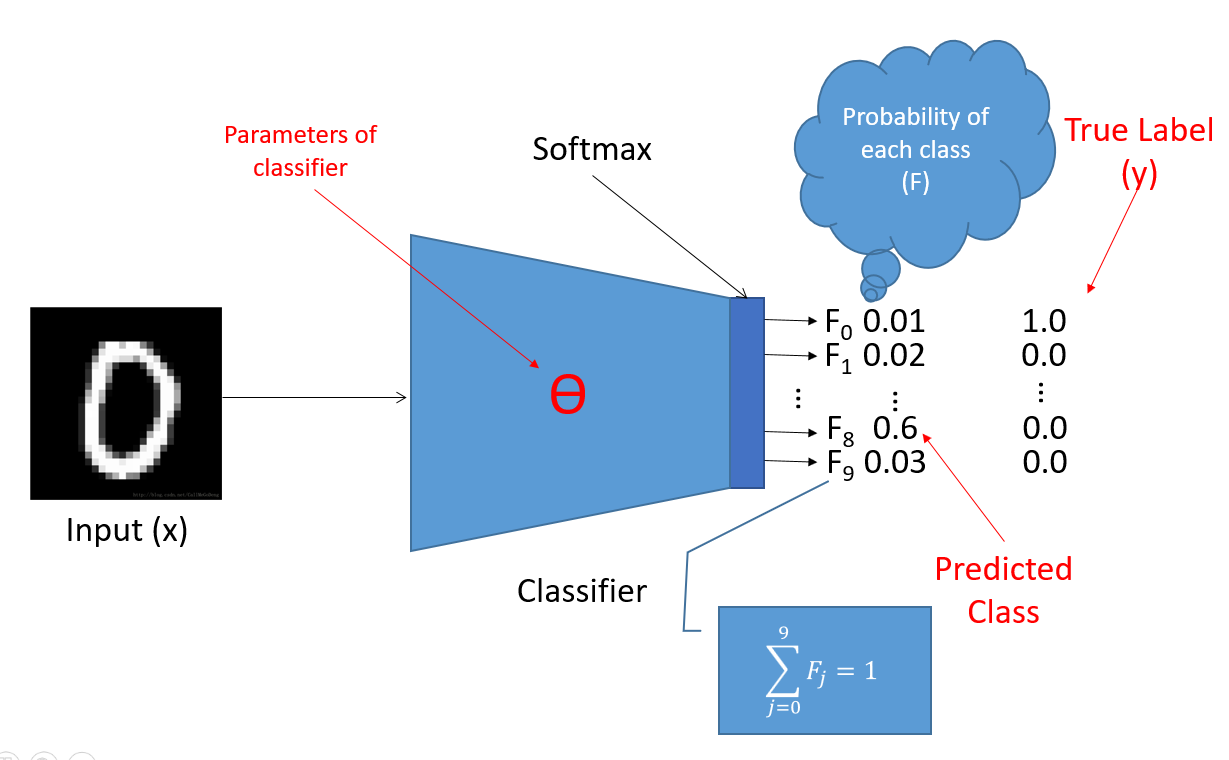}
  \setlength{\belowcaptionskip}{-20pt}

  \caption{Illustration of a DNN classifier.}
  
  \label{fig:dnn_illustration}
\end{figure}

A deep neural network (as a classifier), as illustrated in Figure \ref{fig:dnn_illustration}, is a non-linear function which maps an input to a probability vector where each of its elements \blue{corresponds} to a class score. The element that has the largest score is considered as the prediction. A deep neural network consists of multiple layers 
that are connected to each other sequentially such that the output of one layer becomes the input of the next layer, and each layer applies a non-linear transformation to its inputs. 
Each layer has a set of parameters which are initialized randomly.  
By varying those parameters, the output of the classifier changes. The goal is to find values for the parameters such that for most of the inputs, the neural network predicts their labels correctly, which means that the probability \gr{corresponding} to their true label should be larger than others. 
%
%
%


In order to train this network\gr{,} we want to find the set of parameters that make it predict most of the inputs correctly. So, we have to maximize the score \gr{corresponding} to the true label for each input.
In general, we can say that we need to find a $\theta$ that maximizes $ \sum_{j=0}^K y_{j}F_j$ for every input. Note that $y$ is a vector and only one of its elements is 1 and \gr{the} others are 0. Instead of maximizing $ \sum_{j=0}^K y_{j}F_j$\gr{,} we can minimize $ -\sum_{j=0}^K y_{j}log(F_j)$.  Mathematically, we need to solve the following \gr{optimization} problem to train a model:
\[argmin_\theta \left(-\frac{1}{N}\sum_{i=0}^N \sum_{j=0}^K y_{ij}log(F_j(x_i))\right)\]
\gr{where} N is the total number of samples in our training set and K is the total number of classes. This is called a cross-entropy loss function\gr{,} which is a function of $\theta$. A lower value of this function means better predictions over the training set. \blue{In order to solve this minimization problem, a technique called gradient descent, or one of its variants such as Adam optimization~\cite{adam}, is used.} 


\subsection{Notation}
In the rest of this paper we use the following notation:

\begin{itemize}
  \item $x$: the legitimate (clean) input. $x \in [0,1]^m$, where m is the number of pixels in an image.
  \item $y$: the label corresponding to the legitimate input.
  \item $x^\prime$: the adversarial input. $x^\prime \in [0,1]^m$.
  \item $y^\prime$: the label corresponding to the adversarial input\gr{,} which is different from its original label.
  \item $y_{target}$: the label which an adversary wants to make the classifier output.
  \item $F(.)$: the classifier which maps an image to a label. For correctly predicted inputs we have $F(x) = y$.
  \item $\theta$: the parameters of classifier.
  \item $Z(.)$: the logits\gr{,} which are inputs of the softmax layer. So, $softmax(Z(x)) = F(x)$.
  \item $\delta$: the perturbation which is added to a legitimate example to make it adversarial. So, $x^\prime = x + \delta$.
  
\end{itemize}

  

\subsection{Adversarial Example}
\red{Previous works showed how to craft adversarial examples in white-box and black-box settings \citep{intriguing,cw,fgsm, jsma, deepfool, atn,transferable,practical_bb}.} \gr{ We discuss some of them here.}
\subsubsection{\red{White-box Setting}}
\blue{Early efforts by Szegedy et al. in \cite{intriguing} and Biggio et al. in \cite{biggio_craft} showed how to craft an adversarial example.} The process for crafting a targeted adversarial example can be reduced to a box-constrained optimization problem as follows:\\[-7pt]\par

{\centering

$ argmin_{\delta} ||\delta||_p$ s.t. $ (x+\delta) \in [0,1]^m $ and  $F(x+\delta) = y_{target}$\\[+3pt]\par}

This optimization problem means that an attacker wants to find the minimum perturbation, so that if she adds it to the input, the classifier would predict it as the attacker's desired target. However, neural networks are not convex, so this optimization problem is intractable and people use different heuristics to find a small enough perturbation that can fool the model. 
There is another class of attacks, which are known as non-targeted attacks, in which the attacker's goal is to make the classifier misclassify the input to any other label (as opposed to targeted attacks which the goal is to make the classifier output a specific label).  For non-targeted attacks the optimization problem can be formulated as follows:\\[-7pt]\par
{\centering $ argmin_{\delta}  ||\delta||_p $  s.t. $ (x+\delta) \in [0,1]^m $ and  $F(x+\delta) \neq y$ \\[+3pt]\par}

Carlini et al. in \cite{cw} described a way to craft adversarial examples, and we explain it here briefly as we use the same loss function in our attack.
They designed their attack by introducing a new objective function. The objective function that they used is as follows:\\[-7pt]\par

{\centering minimize $c.||\delta||_p + f(x+\delta)$ s.t. $x+\delta \in [0,1]^m$\\[+3pt]\par}

\noindent\gr{in} which $p$ can be 0,2 or $\infty$.
One of their choices for function f is:\\[-7pt]\par
{\centering $  f(x^\prime) = (max_{i\ne t}(Z(x^\prime)_i) - Z(x^\prime)_t)^+$\\[+3pt]\par}

\noindent\gr{in} which Z is the logit which are the inputs to the softmax function, t is the target label, $(e)^+$ is short-hand for $max(e,0)$, and $c$ is a hyper parameter \gr{that} determines the trade-off between the amount of distortion and the growth of the target score. By decreasing $c$, the amount of distortion and \blue{the} success probability grows. They showed that by using this function, they can craft adversarial examples for the MNIST, CIFAR-10, and ImageNet datasets with less distortion compared to other white-box attacks. This minimization basically says that we want to find a $\delta$ such that its magnitude is minimal (with respect to $l0$, $l2$ or $l\infty$ norm) and the logit value corresponding to the target label is larger than other logits, which makes the classifier predict the input as the target class. This optimization problem is solved by the help of gradient descent, which we mentioned earlier. Carlini and Wagner also showed that they can build adversarial examples \gr{that} will make the classifier output the target label with higher probability by slightly changing the function $f$ as follows:\\[-7pt]\par
{\centering $f(x^\prime) = max(max_{i\ne t}(Z(x^\prime)_i) - Z(x^\prime)_t,-\kappa)$\\[+3pt]\par}

\noindent\gr{in} which $\kappa>=0$ and determines the confidence score. By increasing \blue{ $\kappa$}, the confidence score of the target class becomes larger. This function basically means that we keep modifying the input as long as $ Z(x^\prime)_t < max_{i\ne t}(Z(x^\prime)_i) + \kappa$.

This technique is not the only way to craft adversarial examples. 
For more information about crafting adversarial examples in white-box setting we refer the reader to ~\cite{intriguing, fgsm, jsma, deepfool, atn}.

\subsubsection{\red{Black-box setting}}

Szegedy et al. in \cite{intriguing} \blue{also} showed that, in many cases, an adversarial \gr{example built} using one classifier can fool another classifier that has a different architecture and parameters.  This property is called the transferability of adversarial examples.
\blue{By using this property, Carlini et al. in \cite{cw} showed that they could craft adversarial examples against classifiers fortified by defensive distillation \cite{distillation}, which block gradients by crafting adversarial examples against a different model with high confidence. Later, Liu et al. in \cite{transferable} built on top of this idea and crafted adversarial examples against multiple pre-trained models to then be able to transfer them to the target model. It has been also shown by Tram{\`e}r et al. in \cite{ensemble_adv_train} that augmenting training data with adversarial examples generated by a few fixed, pre-trained models significantly improves the robustness of a model in the face of these types of transferable black-box attacks. }

\blue{
Biggio et al. in \cite{biggio_craft} and Papernot et al. in \cite{practical_bb} showed that an adversary can craft adversarial examples against a model in a black-box setting by querying the target model and training a substitute model using the labels predicted by the target model. In this case, after training the substitute model, the adversary can craft adversarial examples against \gr{the} substitute model in order to transfer them to the target model. \red{Papernot et al. in \cite{practical_bb} also showed independently that they can evade defensive distillation by querying the target model in a black-box setting.}  }

\blue{
Our work is built on top of this transferability property and substitute training approach and provides a better tool for evaluating defenses by considering a more powerful attacker \gr{that} has access to the logits of the target model.} 

\subsection{Defenses}

In general\gr{,} there are two different approaches for defending against adversarial examples:

\begin{itemize}
\item Fortifying Defenses: These types of defenses try to make the classifier predict adversarial examples as their correct class. \blue{Techniques \gr{that} are used 
include removing adversarial perturbation by transforming the input before feeding it to the classifier, quantization or discretization of the input, and randomization of the input or the model.}
\item Detecting Defenses: For these types of defenses, the classifier may predict an adversarial example incorrectly, but there is an adversarial detection mechanism \gr{that} makes the whole model reject those cases. A \gr{technique used} for these defenses is to augment the classifier with another DNN (or any other model) to classify the input as legitimate or adversarial, or other means of detection such as using
some statistics which are assumed to be co-related with adversarial examples.
\end{itemize}

\section{Stochastic Substitute Training}

\red{In this section, we introduce our new gray-box approach to generating adversarial examples.}

\begin{figure}
  \centering
  \includegraphics[width=65mm]{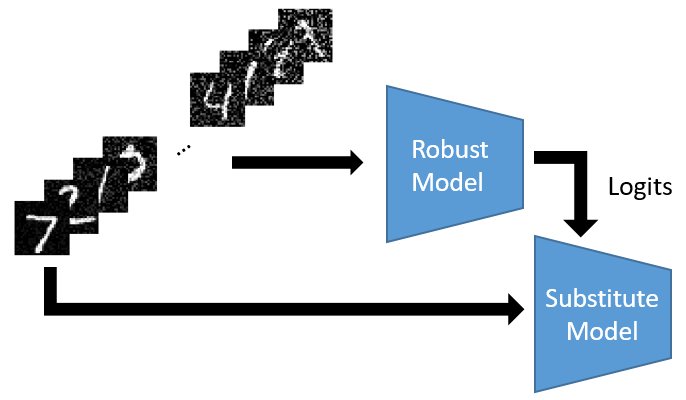}
  \setlength{\belowcaptionskip}{-7pt}
  \caption{Illustration of Stochastic Substitute Training.}
  \label{fig:rand_noise}
  
\end{figure}

\red{\subsection{Threat Model}}
Before describing our approach, it is useful to clarify the threat model.  In this paper we consider two different threat models:
\begin{itemize}
\item For evaluating fortifying defenses, we consider an attacker that can send inputs to the model and see the logits. The attacker is not aware that a defense is in place and she doesn't have access to the model or defense parameters.
\item For evaluating the defenses that detect adversarial examples, we consider an attacker that knows a detection mechanism is in place. She can send inputs to the model and see logits and the output of the detector but doesn't have access to the model or detector parameters.
\end{itemize}
\red{\subsection{Algorithm}}
In order to attack the robust classifiers with defenses that obfuscate gradients, we use the transferability property of adversarial examples. In general, we add  \blue{different levels of random noise} to the set of images we want to craft adversarial examples for.  In the case of our experiments, this would be the test set of MNIST and CIFAR-10. We then feed this dataset to the robust classifier and record the logits. After that, we train a substitute model with this dataset and the recorded logits. Figure \ref{fig:rand_noise} illustrates this process.

For training this model, instead of using a default cross entropy loss, we use the mean square error between the substitute model's logits and logits we got from \gr{the} robust model as our loss function. \gr{More specifically, the loss function is defined as follows:}

\[ Loss_{SST} = \frac{1}{N}\sum_{i=0}^N \sum_{j=0}^K \frac{1}{K}\left( Z_j^{robust}(x_i+r_i) - Z_j^{sub}(x_i+r_i) \right)^2  \]

\noindent \gr{where} $r_i$ is the \red{noise} added to the sample $x_i$ and $N$ is the total number of inputs in our augmented dataset.  Training a substitute model in this way makes the substitute model's decision boundaries for that dataset very close to the robust model, which makes transferability to the robust
model easier.  Further, since the substitute model is differentiable, we can craft adversarial examples for it using an iterative method. Training a substitute model with images augmented with random \red{noise} helps the substitute learn how the robust model's class probabilities change in the neighborhood of each sample. Note that these types of random \red{noises} do not necessarily change the prediction of the robust model, but it helps the substitute detect in which directions the correct class score can be decreased. Also, because we assumed that the attacker doesn't know the defense which is in place and how robust the model is, we augment the dataset with \blue{different levels of \red{random noise}}. This is because if we add a low level of \red{random noise}, the model might be very robust and 
\red{the adversarial perturbations found during the crafting procedure} may not be sufficient to fool the classifier.  On the other hand, if we add a high level of \red{random noise}, the model may not be that robust and unnecessary \red{adversarial perturbations} would be added to the \red{images during crafting procedure}. The reason we use logits instead of class probabilities is that the effect of low level \red{random noise} is more obvious in logits compared to probabilities. Using probabilities may not capture the impact of small amount of random \red{noise} because of \blue{the} floating point precision. We empirically found that for complex datasets, training multiple copies of a model with different random \blue{noises} reduces the required \blue{adversarial} perturbation on average. We speculate this is because the decision boundaries of the robust model and substitute models are not completely matched, and each of the substitute models \red{approximates} the decision boundaries for some specific images better than others.

After training the substitute model for multiple epochs, we craft adversarial examples against it by using the C\&W loss function \gr{mentioned} in Section 2.3 and gradient descent to minimize this loss function iteratively.  By doing so, we find a small adversarial perturbation that fools the robust model. We minimize the loss function by using the Adam algorithm as our optimizer. In each iteration, we check if the current perturbation can fool the robust model. If so, we increase the $c$ parameter in \gr{the} C\&W loss function to craft adversarial examples with a smaller amount of distortion in subsequent iterations. We also keep decreasing the value of $c$ in each iteration \blue{until we} fool the robust model in order to increase the amount of perturbation and chance of transferability. If we couldn't find an adversarial example in the first run, we restart the algorithm and increase  $\kappa$ to build adversarial examples with higher confidence. This might increase the amount of perturbation, but it also increases the chance of transferability to the robust model. Algorithm 1 shows this process for crafting adversarial examples. In this procedure $F$ is the target classifier.

\red{\subsection{Benefit of Noisy Data Augmentation}}

\blue{In order to show the benefit of our stochastic substitute training over an approach \gr{that} uses a substitute model \gr{trained} without data augmentation, we first trained a model on MNIST as our target model. Then, we trained a substitute model with and without augmenting data with random noises with the first 100 samples from MNIST test set. We made 2100 replication of this set and trained the substitute model on this dataset one epoch with lr=0.001 and another epoch with lr=0.0001. Different levels of random \red{noises} were added to different replicas for the one used in SST. Then with each of the substitute models, we crafted an adversarial example with Algorithm 1 for the first 100 samples. The average l2 norm of adversarial examples crafted with stochastic substitute training was 1.57.  The average l2 norm for the substitute model which was trained without data augmentation was 3.10. 
As it can be seen, the average perturbation of images crafted without SST is almost 2 times larger than those which are crafted with SST.}

\blue{
For the sake of comparison, 
to measure what we are sacrificing by limiting our approach to a gray-box setting, 
we also crafted adversarial examples with the C\&W attack for the target model in a white-box setting. The average l2 norm of adversarial examples crafted in this way was 1.25, which can be seen as the minimum required perturbation found by current white box techniques to fool the target model for those images. As it can be seen, crafted adversarial examples with C\&W are only slightly better than those crafted with SST.}
\begin{algorithm}
\caption{Crating adversarial examples}\label{alg1}
\begin{algorithmic}[1]
\Procedure{CraftAdvExample}{$x,totalRun,totalIter,F$}
\State $adv \gets [0]^m$ \#Adversarial Example
\For{\texttt{ each \textit{i} $\in [0,totalRun]$}}

	\State initialize $\delta$ randomly 
    \For{\texttt{ each \textit{j} $\in [0,totalIter]$}}
		\State take one step of GD using Adam
        \State $x^\prime \gets Clip(x+\delta)$
        \If {$adv == [0]^m$}
        \State decrease $c$
        \EndIf
 		\State \# Check detector's prediction as well (if any)
        \If {$F(x^\prime)\neq y$ and $||x-x^\prime||_2 < ||x - adv||_2$ }
        
        \State $adv \gets x^\prime$
        \State increase $c$
        \EndIf

    \EndFor
    \State increase $\kappa$

\EndFor
\EndProcedure
\end{algorithmic}
\end{algorithm}
\section{Evaluation of SST against Fortifying Defenses}

In this section we evaluate the effectiveness of SST
against two fortifying defenses --
\blue{random feature nullification} (RFN) \cite{rfn} and \blue{thermometer encoding} \cite{therm}.

\subsection{Random Feature Nullification}
Wang et al. in \cite{rfn} proposed an adversary resistant technique to obstruct attackers from constructing impactful adversarial samples.  They called this adversarial resistant technique ``random feature nullification'' and is described as follows:

For each batch of inputs denoted by $X \in R^{n \times m}$, where $n$ is the number of samples and $m$ is the feature vector size, random feature nullification performs element-wise multiplication of $X$ with a randomly generated mask matrix $I_p \in R^{n \times m}$, where its elements are only 1 or 0. The result is then fed to the classifier. During training they generate a mask matrix in a way to randomly select the number of features to nullify and also randomly select which features to nullify. More specifically, for each row of $I_p$ denoted by $I_{p^i}$, the number of features to be nullified are selected from the Gaussian distribution $N(\mu_p,\sigma^2_p)$, and then a uniform distribution is used for generating each row of ${I_p}$. During test-time, the nullification rate is fixed to $\mu_p$, but choosing features in each sample is still random with uniform distribution. The randomness they introduced during test-time prevents an adversary from computing the gradients needed for crafting an adversarial example. In their evaluation they showed that a classifier fortified by RFN can resist against 71.44\% of generated adversarial examples in the case where the adversary is allowed to change the value of each pixel by 0.25.

\subsubsection{Our Evaluation}

Since the authors didn't publish their code, in order to evaluate RFN, we trained a model with the same architecture and parameters they used in their paper.  More specifically, the parameters can be found in Table \ref{tab:rfn_params} in \blue{the Appendix}. For the hyper parameters of RFN, we set  $\mu_p$ to 0.5 and $ \sigma $ to 0.05. During test time for each input, half of its features are nullified before feeding to the DNN.
After training the model, we got 96.63\% accuracy on the MNIST test set.

For training the substitute model, we added uniform random \red{noise} to the test set and created a new data set with 70000 samples. For the first 10000 samples the \red{range of noises} was in $[-0.05,0.05]$, for the next 10000 samples \blue{it} was in $[-0.1,0.1]$, and so on.
We used Adam optimizer to train the substitute model with 0.001 learning rate for 10 epochs and then 5 epochs with 0.0001 learning rate. It finally reached 97.91\% accuracy on legitimate test samples. The substitute model architecture can be found in Table \ref{tab:sub_rfn_params} \blue{in the Appendix}. In this table, the convolution layer parameters are described as $M,K\times K,S$ which refers to a convolution layer with M feature maps, filter size $K \times K$ and stride S. The Max Pooling layer parameters are described as  $K\times K,S$, which refers to a Max Pooling layer with pool size $K \times K$ and stride S.

\begin{figure}
  \centering
  \includegraphics[width=45mm]{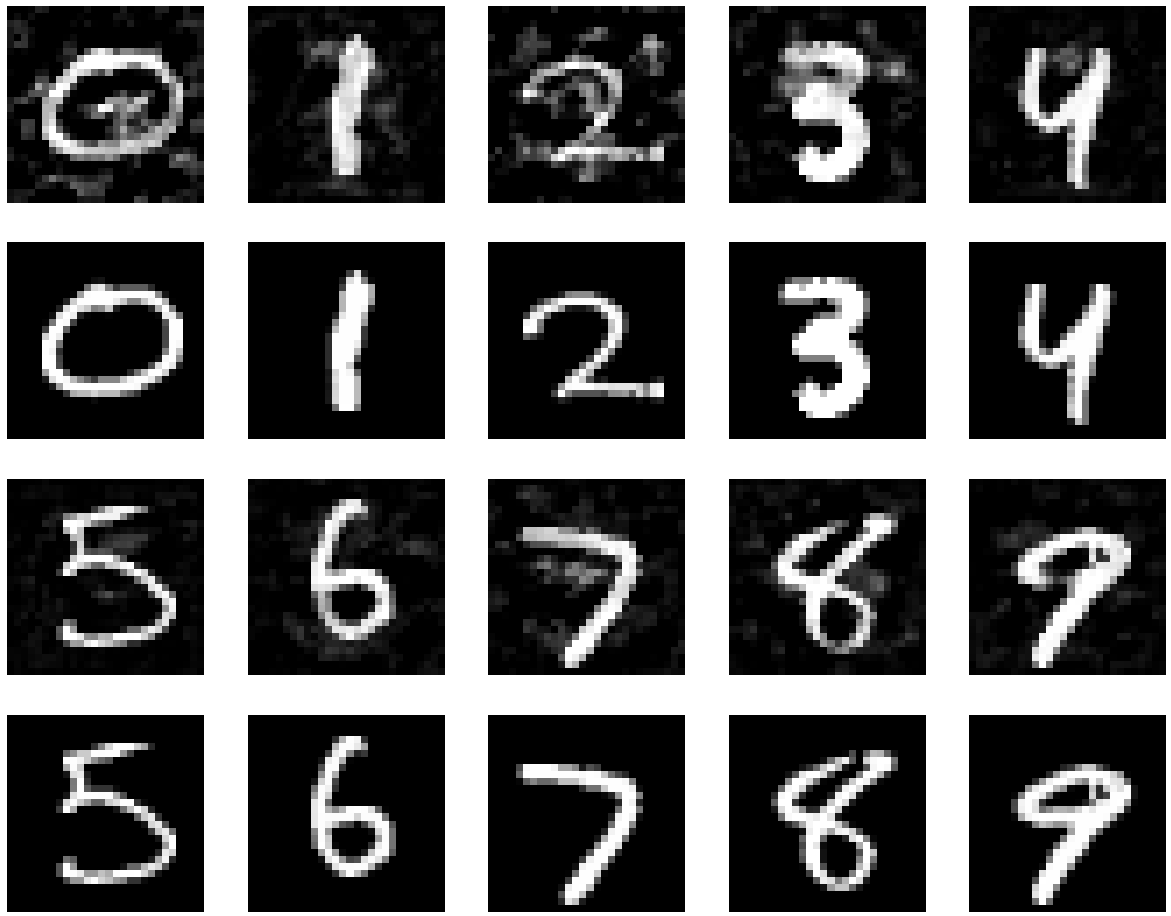}
  \setlength{\belowcaptionskip}{-15pt}
  \caption{Adversarial examples crafted for the MNIST dataset against a classifier fortified with RFN.  They are misclassified for more than 50 times in 100 parallel runs.}
  \label{fig:rfn_samples}
\end{figure}

Since RFN is a stochastic defense, feeding the same image to the model multiple times may cause different results. So, an adversarial example may fool the classifier in one run, but it may be predicted correctly in the next run. In the paper, it is not discussed what exactly should be considered as fooling the model. Here we report the accuracy of \blue{the} model on legitimate samples and the average l2 norm that is required to fool this model in different scenarios. First, we consider an input to be classified correctly if in 100 parallel \gr{runs} the model can predict it correctly for more than 50 cases. In the second scenario, we change this threshold to 70, and for the last one we change it to 90. For the evaluation of the model, we used the first 100 samples in the test set to generate adversarial examples using our attack. For crafting adversarial examples, we set the learning rate to 0.001. For each sample, we chose the target label as the second most probable class predicted by \gr{the} robust model. In \gr{Algorithm 1}, we set the total run to 3 and total iterations to 300.  First, we evaluated this model with an $l_2$ attack. The average $l_2$ norm for different scenarios can be found in Table \ref{tab:rfn_results}. The success rate in all cases was 100\%. Figure \ref{fig:rfn_samples} also shows the crafted adversarial examples against this model when the threshold is 50. 

For this defense we also generated adversarial examples with $l_\infty=0.25$.
We could generate adversarial examples for 94 samples out of 100 in the first scenario. So, in this case, the resistance rate is only 6\%. In the second scenario, we could generate adversarial examples for 92 samples. For the last scenario, we could generate adversarial examples for 75 samples.


\begin{table}
\centering
    \begin{tabular}{ |>{\centering\arraybackslash}m{2.0cm} | >{\centering\arraybackslash}m{2.0cm}| >{\centering\arraybackslash}m{2.0cm}|}
    \hline
    Threshold & Accuracy & L2 norm \\ \hline
    50 & 97.93 & 2.13 \\ \hline
    70 & 96.46 & 2.43 \\ \hline
    90 & 92.78 & 2.96 \\ \hline

    \end{tabular}
    \caption{The accuracy of a model fortified with RFN and the average l2 norm of the required perturbation across different thresholds.} \label{tab:rfn_results}
\end{table}


\subsection{Thermometer Encoding}

Buckman et al. in \cite{therm} introduced another defense, called thermometer encoding, \gr{which} \blue{makes a model more robust} against adversarial examples.  It prevents an adversary from calculating the gradients \gr{that} are needed for crafting adversarial examples. Thermometer encoding is applied to each pixel of the input before feeding it to the classifier to discretize it. The way it works is as follows: for each pixel $p$ of \gr{the} image the k-level thermometer-encoding $\tau(p)$ is a k-dimensional vector where \par


{\centering $    \tau(p)_j= 
\begin{cases}
    1,& \text{if } p\geq j/k\\
    0,              & \text{otherwise}
\end{cases}$\par}

and $\tau(p)_j$ is the j-th element of the vector. For example\gr{,} for a 10-level thermometer encoding, $\tau(0.33) = 1110000000$. Since this function does a discrete transformation, it is not possible to back-propagate gradients through it. \gr{Therefore}, an adversary can't craft adversarial examples for it using traditional white-box techniques.

\subsubsection{Our Evaluation}

For evaluating the effectiveness of SST against this defense, we used the model trained by Athalye et al. in \cite{obfuscated}, which is a wide ResNet model \cite{wideResNet} fortified by thermometer encoding trained on CIFAR-10. For training this model, the adversarial training technique introduced by Madry et al. in \cite{madry} is also used for more robustness.

For attacking this model, we trained multiple substitute models with different levels of random \red{noise}. The model architecture we used as our substitute model is described in Table \ref{tab:therm_sub} \blue{in the Appendix}. We trained four copies of this model on the CIFAR-10 test set, which we refer to as A, B, C and D. More specifically, we created a new dataset by replicating the CIFAR-10 \blue{test set} eight times and adding \blue{different levels of random noises} to it.  We trained each substitute model with the training procedure we described in Section 3.
\blue{The range of noises we added for training model A was $[-\frac{2}{255} \times i,\frac{2}{255} \times i]$ for $i \in [1,8] $, where $i$ was incremented for each replica. This range for Model B, C and D was $[-\frac{3}{255} \times i,\frac{3}{255} \times i]$, $[-\frac{4}{255} \times i,\frac{4}{255} \times i]$, and $[-\frac{5}{255} \times i,\frac{5}{255} \times i]$ respectively.}
Each of the substitute models was trained by the Adam optimizer as follows: 6 epochs with lr=0.001, 3 epochs with lr=0.0005, 3 epochs with lr=0.0001, 3 epochs with lr=0.00005, 3 epochs with lr=0.00001, 3 epochs with lr=0.000005, and  3 epochs with lr=0.000001.

\begin{figure}
  \centering
  \includegraphics[width=50mm]{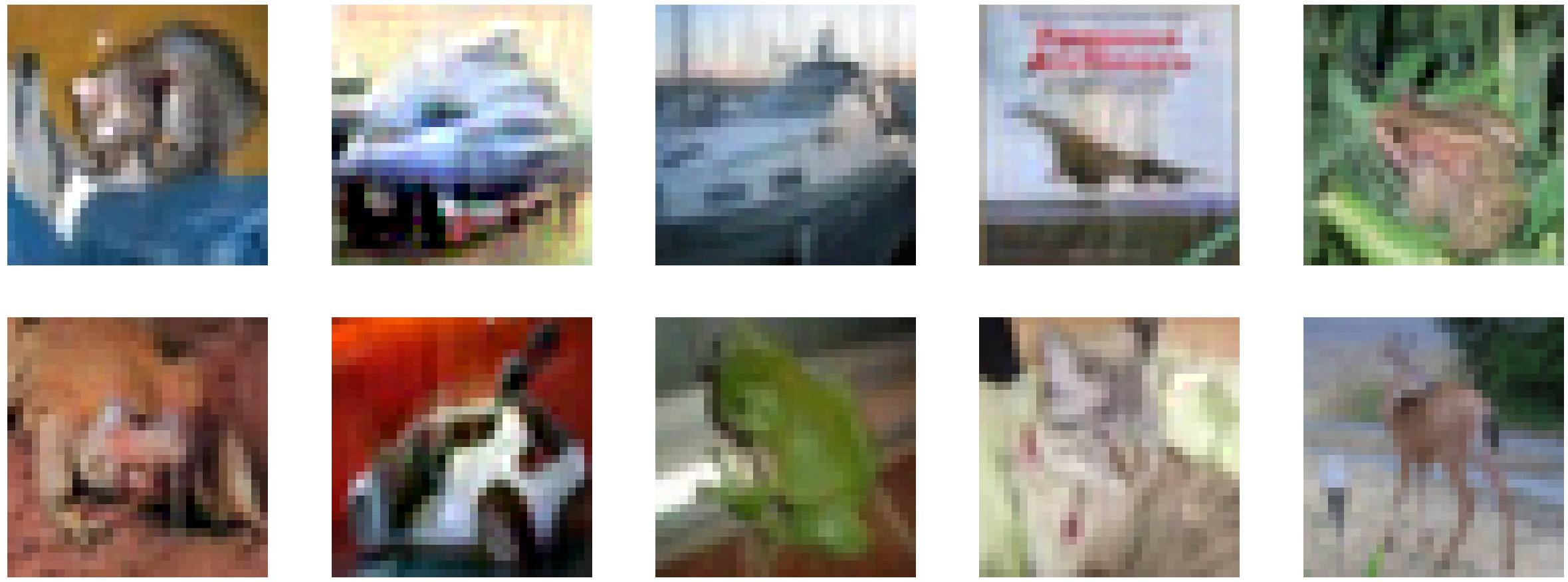}
  \setlength{\belowcaptionskip}{-15pt}
  \caption{Adversarial examples crafted for the CIFAR-10 dataset against a classifier fortified with thermometer encoding using 4 substitute models.}
  \label{fig:therm_samples}
\end{figure}

We finally crafted adversarial examples using these models.  For the first 100 images in the CIFAR-10 test set \gr{that} were predicted correctly by the robust model, we set the total run to 3 and total iterations to 600. After every 100 iterations, we restarted the perturbation randomly to reduce the impact of sticking in a local minimum. Table \ref{tab:therm_results} shows the success rate and average l2 norm for different scenarios in addition to average time for crafting one adversarial example. Figure \ref{fig:therm_samples} shows the adversarial examples generated using all 4 models.
\red{The reason that we couldn't find an adversarial example in some cases is that our substitute models didn't approximate the decision boundaries of the target model well enough in those cases. After some iterations, because of the values we chose for $\kappa$, $ Z^{sub}(x^\prime)_t$ becomes greater than $max_{i\ne t}(Z^{sub}(x^\prime)_i) + \kappa$. As a result, the loss function becomes a constant value, \gr{with a gradient $0$.} Thus, newer perturbations won't be added to the current perturbation and the attack doesn't progress. We speculate that this problem can be solved by choosing a higher \gr{value} for $\kappa$ for those cases \gr{where the} attack fails. The cost is a higher level of perturbation.}

\begin{table}
\centering
    \begin{tabular}{ |>{\centering\arraybackslash}m{1.25cm} | >{\centering\arraybackslash}m{1.25cm}| >{\centering\arraybackslash}m{1.25cm}| >{\centering\arraybackslash}m{1.25cm}|}
    \hline
    Substitute Model(s) & Success Rate & L2 Norm & Time \\ \hline
    A,B,C,D & 100\% & 2.79 & 69 sec \\ \hline
    A,B & 99\% & 3.14 & 58 sec \\ \hline
    C,D & 99\% & 2.96 & 56 sec \\ \hline
    A & 96\% & 3.46 & 51 sec \\ \hline
    B & 99\% & 3.52 & 51 sec \\ \hline
    C & 97\% & 3.45 & 51 sec \\ \hline
    D & 99\% & 3.54 & 51 sec \\ \hline


    \end{tabular}
      \setlength{\abovecaptionskip}{-5pt}
    \caption{Results of applying our attack against thermometer encoding.} \label{tab:therm_results}
\end{table}

\section{Evaluation of SST against Detecting Defenses}

In this section we evaluate the effectiveness of SST against the SafetyNet \cite{safetynet} and Defense-GAN \cite{defensegan} detecting defenses.

\subsection{SafetyNet}

Metzen et al. in \cite{detector_network} introduced a way to detect adversarial examples by augmenting the classifier with another DNN which acts as a detector. This detector network is trained with the outputs of some intermediate layer of \gr{the} original classifier while adversarial examples and legitimate examples are fed into it.  Later, Carlini et al. in \cite{carlini_detection} showed that an adversary can craft adversarial examples against this defense by changing the loss function such that an adversarial example can be crafted by back-propagating through both the original classifier and detector.

But, Lu et al. in \cite{safetynet} introduced another mechanism for adding a detector 
called SafetyNet. In SafetyNet, the detector is still connected to the output of some late layer of the classifier. But they used two techniques which \gr{make} it impossible for the adversary to get any gradient from the detector. The first technique is quantization, in which the outputs of ReLU is quantized at some specific thresholds. The other technique is using a support vector machine (SVM) with a radial basis function (RBF) kernel as the detector.  This provides no useful gradient to the adversary. They also observed that ``there is a trade-off between classification confidence and detection easiness for adversarial examples. Adversarial examples with high confidence in wrong classification labels tend to have more abnormal activation patterns, so they are easier to be detected by detectors.'' As a result, the classification confidence is also considered in SafetyNet. For each input, the ratio of \gr{the} second highest classification confidence to the highest classification confidence is calculated, and if it is bigger than a specific threshold that example is rejected. For our experiments, we set this threshold to 0.25, as suggested in the original paper.

\subsubsection{Our Evaluation} 

Since the code for SafetyNet was not published, we implemented their defense ourselves on a model trained on MNIST.
The model architecture we used for training is described in Table \ref{tab:safeteynet_params} in \blue{the} Appendix. We trained this model with the Adam optimizer: 3 epochs with lr=0.001 and 3 epochs with lr=0.0001. The accuracy of this model on the MNIST test set was 99.15\%, and the average confidence of correctly classified images was 99.63\%.

To train the detector, we first generated non-targeted adversarial examples for the first 5000 samples of the training set using the C\&W attack. For training the detector, we used the outputs of the first fully connected layer (layer 4), and we quantized them into four bins before feeding \blue{them} to the SVM with the RBF kernel. Since the SafetyNet paper does not describe how the thresholds for quantization should be chosen, we chose them as follows: We fed our training data to the classifier and collected the outputs of layer 4. We sorted all the positive values from this data and found the 1st quartile $Q_1$, median, and 3rd quartile $Q_3$ and used them as thresholds for quantization. So all of the 0s, and any value less than the 1st quartile, were converted to the middle of that bin (i.e. $\frac{(Q_1 - 0)}{2}$). Any value between the 1st quartile and median was converted to the middle of the second bin (\textit{i.e.,} $Q_1 + \frac{median - Q1}{2}$), and so on. Note that our attack works regardless of the way quantization thresholds are chosen. We decided to choose thresholds in this way as it gave us a good accuracy for training the detector. After training the detector in this way, it could achieve 95.15\% accuracy on the MNIST test set.

In order to attack this defense, we trained two models with the logits of the classifier in the same way we did for RFN. We also trained another model with probability scores of the detector while the same data set was fed into the classifier. The architecture of the substitute classifier was same as the one we used for attacking RFN. The architecture of the substitute detector can be found in Table \ref{tab:safeteynet_sub_detector} \blue{in the Appendix}. Note that we assumed that the attacker doesn't know where the detector is connected or what the input to it is. So, the substitute detector is trained on raw pixel values. 

\begin{figure}
  \centering
  \includegraphics[width=45mm]{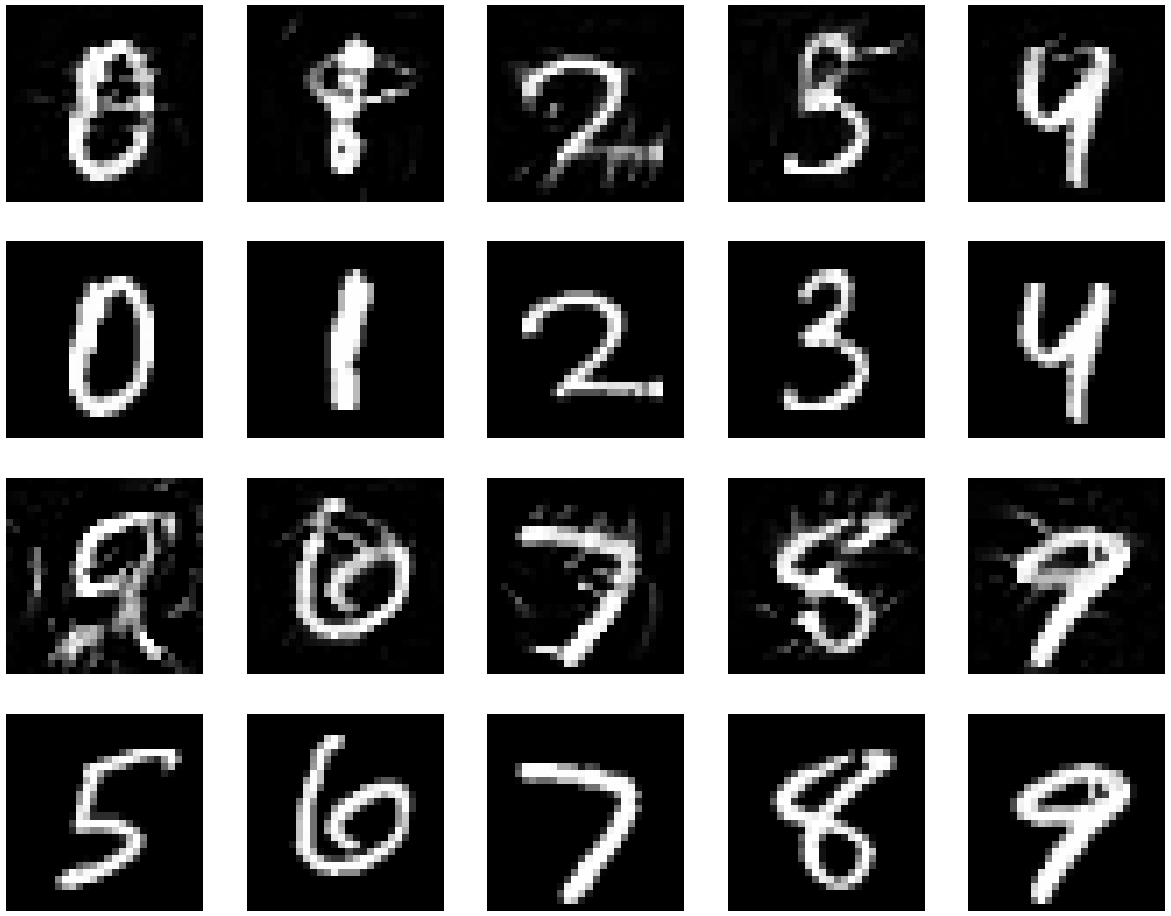}
  \setlength{\belowcaptionskip}{-19pt}
  \caption{Adversarial examples crafted against SafetyNet for the MNIST dataset.}
  \label{fig:safetynet_adv_images}
\end{figure}

With the procedure we introduced in Algorithm 1, we crafted adversarial examples against SafetyNet. We set the total run value to 3 and total iterations to 300. The only difference was that in the inner loop we checked three things to make sure that the defense is bypassed. First, we checked to make sure that the adversarial example fools the classifier. Second, we checked that the confidence ratio is less than 0.25, and finally we checked to make sure the detector predicts it as a legitimate sample. We crafted adversarial examples for the first 100 samples that are classified correctly and predicted as legitimate samples by the detector. The average l2 norm of crafted adversarial examples was 3.37, and the success rate was 98\%. You can see a few samples in Figure \ref{fig:safetynet_adv_images}. In this figure the adversarial examples are in \blue{the} first and third rows and corresponding legitimate examples are in the second and forth rows.

\subsection{Defense-GAN}

Samangouei et al. in \cite{defensegan} introduced a defense \gr{that} makes a classifier more robust against adversarial examples.  They also provided a mechanism to detect adversarial examples in case an attacker \blue{could} fool the classifier. They called this defense, Defense-GAN as they used a Generative Adversarial Network (GAN) as part of their defense. A GAN consists of a generator $G$ and a discriminator $D$ in which, given a random vector $z$ \gr{the} generator tries to generate samples from a particular distribution and the discriminator tries to discriminate between a real sample and a generated sample by $G$. After training a GAN, the generator learns to generate a sample from that distribution. It is also shown that given an image $x$ from that distribution, a vector $z$ can be found such that the reconstruction error $||G(z) - x||_2 \longrightarrow 0$. So, this vector can be found by choosing a random vector $z$ and then minimizing $||G(z) - x||_2$ with gradient descent for $L$ steps and updating $z$ in each step. In their defense, a GAN is first trained on legitimate examples. 

Then at test time, given an image $x$ and a random vector $z$, they first project $x$ into the range of \gr{the} generator using $L$ steps of GD to minimize $||G(z) - x||_2^2$, and they do it $R$ times for $R$ different random vector $z_1, z_2,...z_R$. Then, they choose the vector $z^*$ that has a smaller reconstruction error and feed $G(z^*)$ to the classifier. They argued that by using this technique, a lot of adversarial \blue{perturbations} will be removed from an image before feeding it to the classifier.
Also, intuitively, given an image $x$ after $L$ iterations of GD, the reconstruction error is small if $x$ was a legitimate example and large if it is otherwise. They used this property as their detection mechanism. So, given a threshold $\theta$ if $||G(z^*) - x||_2^2$ is larger than $\theta$ it is considered as \blue{an} adversarial example and otherwise legitimate.  

\subsubsection{Our Evaluation} 

In order to evaluate the effectiveness of SST against \blue{Defense-GAN}, we first trained a GAN with their published code for 200,000 iterations. During the test time, we set $L=200$ and $R=10$.

The model architecture and parameters we used for the GAN's generator and discriminator are \gr{the} same as those \gr{that} were used in \cite{defensegan}. For the classifier, we used the model described in Table \ref{tab:model1} \blue{in the Appendix}. After training, the accuracy of this model on the MNIST test set was 99.26\%. The accuracy on the output of \gr{the} generator was 97.13\%. Since it is not specified in the paper how $\theta$ should be chosen, we chose it in a way \gr{such} that all correctly classified legitimate images are considered legitimate. Based on 9713 samples that were classified correctly, we set $\theta = 0.0536$, which is the largest reconstruction error for a correctly classified sample.

The architecture we used for the substitute \blue{model} is the same as the one we used for attacking RFN. The dataset we used for training the substitute model was 7 replications of the MNIST test set, where a specific amount of random \red{noise} was added for each replication. The range of random \red{noise} we added was $[-0.15 \times i,0.15 \times i]$ for $i \in [1,7], $ where $i$ was incremented for each \blue{replica}. We trained four substitute models with this dataset. Crafting adversarial examples with this approach is very slow because querying \blue{Defense-GAN} takes a long time. 
This makes training substitute models very slow, and in the process of crafting adversarial examples, we have to query \blue{Defense-GAN} again in each iteration. 
We decided to craft adversarial examples with this approach for 15 samples. The average time was 156 seconds and success rate was 80\%, and the average l2 norm of successful adversarial examples was 4.00. 
\red{We speculate that our attack is less successful against this defense 
because of the generator. The generator removes a lot of random noise from our augmented dataset before feeding \gr{it} to the classifier. In other words, it maps several distant inputs in our dataset to very close points before feeding them to the classifier. This behavior results in a smaller amount of variation in logits and makes SST less effective. One way to reduce this impact is to query the target model for a larger dataset, but this would result in a process that is very slow.}

\begin{figure}
  \centering
  \includegraphics[width=45mm]{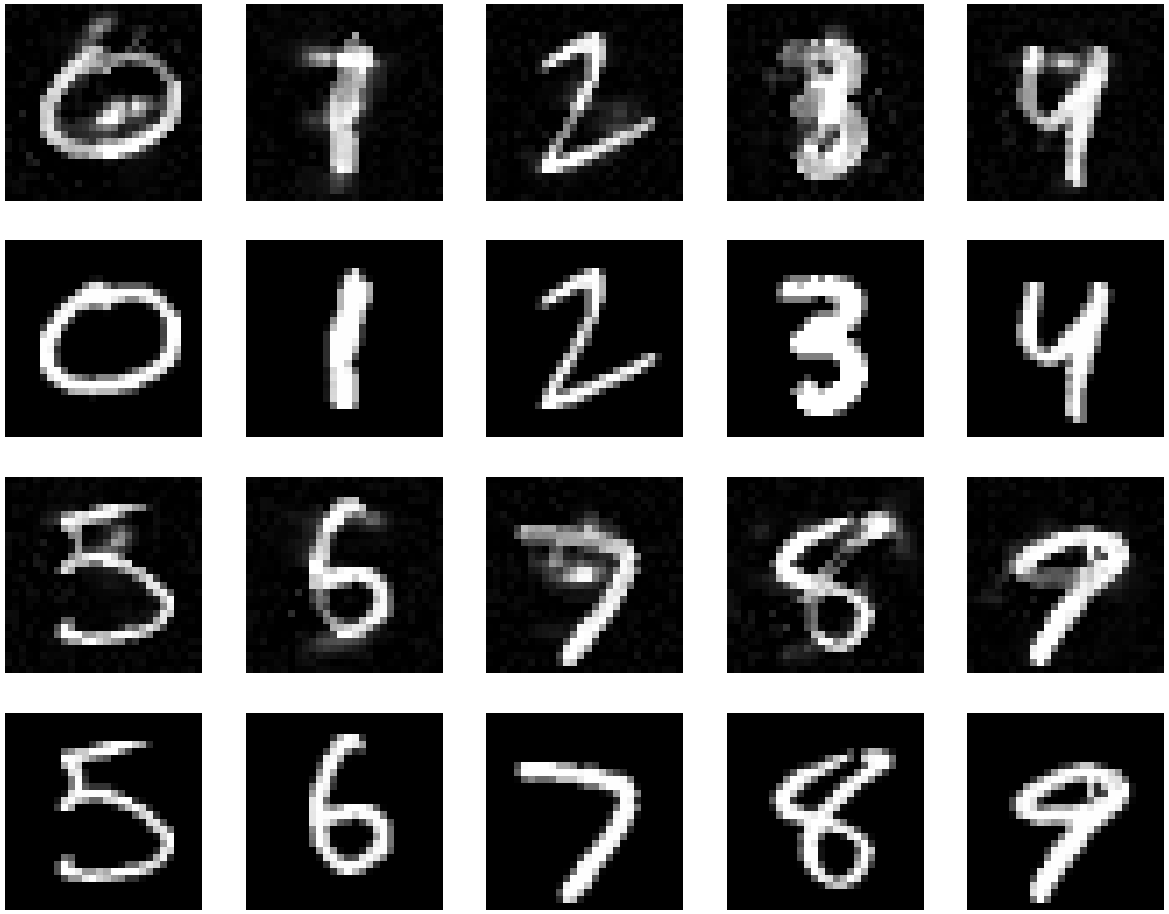}
  \setlength{\belowcaptionskip}{-15pt}
  \caption{Adversarial examples crafted against Defense-GAN for the MNIST dataset.}
  \label{fig:dgan_adv_images}
\end{figure}

In our second attack, we trained a substitute model on the MNIST test set plus random \red{noise} with \gr{a} range $[-0.95,0.95]$ for 20 epochs, and in each \blue{epoch} we updated the random \red{noise}. We didn't query \blue{Defense-GAN} for training this model and used the default class labels and cross entropy loss for training. In order to make the crafting process faster in each iteration, we first \blue{checked} whether the current adversarial example can fool a \blue{model augmented by Defense-GAN}  with $L=30$ and $R=1$.  If we could fool it, we then \blue{checked} it against the model with default parameters ($L=200$ and $R=10$).
Since \blue{Defense-GAN} has stochastic behavior for the same input in different runs, the output might be different. Thus, it is not clear when we should consider an attack successful. So, we \blue{considered} an attack successful if the crafted adversarial example could fool the \blue{Defense-GAN} in three consecutive runs. The sample adversarial examples crafted against this defense can be found in Figure \ref{fig:dgan_adv_images}. The samples are the result of running the attack while setting the target class to the second, third, and forth most probable classes, scored by the original model and choosing the least perturbed one. The success rate for the first 100 samples in the test set, which were classified correctly by \blue{the} robust classifier, was 100\%, and the average l2 norm of perturbations was 3.23. The maximum reconstruction error was 0.0431, which is less than $\theta$.  Therefore, all of the crafted adversarial examples are considered as legitimate examples by the detector. The average reconstruction error was 0.0157. Note that this attack is even more powerful than the first approach, as the attacker doesn't know anything about the attack, the detection mechanism, or the classifier's parameters.




\section{Discussion of Related Works}

\blue{In this section, we evaluate the aforementioned defenses with two types of black-box attacks (mentioned in Section 2.3) that can also be used to craft adversarial examples with no knowledge about a defense and no need to be tailored towards different defenses. While black-box attacks might be more practical than the gray-box setting we considered, as our experiments in this section show, they are not good for evaluating new defenses, such as those we considered in this paper, as in many cases they can't find \gr{an} adversarial example or the required perturbation is so high that it makes it hard even for a human to label correctly. }

\subsection{Black-Box Attack with Jacobian based Dataset Augmentation}

Papernot et al. in \cite{practical_bb} introduced a black-box attack that many researchers have used to show the robustness of their defense in a black-box setting. 
Similar to our attack, the authors in \cite{practical_bb} showed how to attack a 
model using a \blue{small} synthetic dataset without having access to \gr{the} DNN's parameters or knowing about the defense \gr{that} is in place. Their attack was of the non-targeted attack type in that they only made the 
model to mis-classify the inputs. \blue{The threat model they considered was different from what we considered in this paper.} They assumed that an attacker only can send input to the target model and observe its predicted class, where we assume the the attacker also has access to the logits.

To attack the target model, they trained a substitute model to approximate the target model decision boundaries through a procedure called Jacobian-based Dataset Augmentation. 
The way it works is as follows: First they collect an initial small dataset. Then, they label this dataset using an Oracle (black box) model. Next, they train a substitute model using this dataset. Then, for each input $x$ in their dataset, they evaluate the sign of the Jacobian matrix dimension corresponding to the label assigned to $x$ by the oracle: $sgn(J_F(x)[O(x)])$, where $F$ is the substitute model and $O(x)$ is the label assigned to the input by oracle. They then augment their initial dataset with new points created as follows: $x_{new} = x + \lambda.sgn(J_F(x)[O(x)])$, where $ \lambda $ is a hyper parameter. Finally, they repeat these steps for a few iterations (substitute training epochs) so the substitute model can approximate the oracle decision boundaries.

After training a substitute model, 
the attacker crafts adversarial examples for the substitute model with the help of the JSMA and FGSM approaches in order to transfer them to the black-box model.
We implemented this attack against the four defenses we considered, and the results can be found in Table \ref{tab:bb_results}. 
In all the cases, we \gr{crafted} adversarial examples with the FGSM attack for the substitute model and checked what percentage of them can fool the robust model. 

\begin{figure}
  \centering
  \includegraphics[width=50mm]{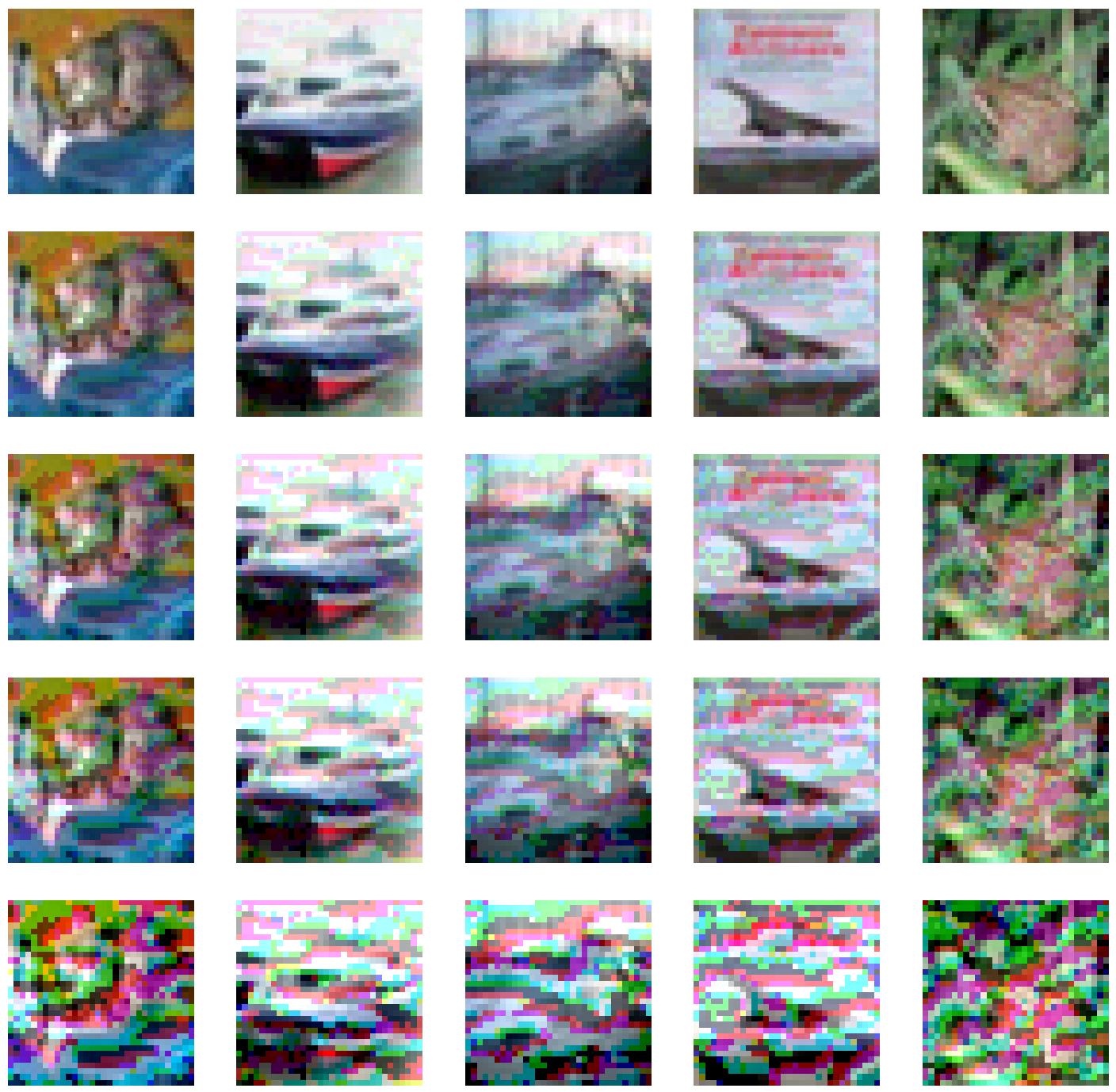}
  \setlength{\belowcaptionskip}{-10pt}
  \caption{\red{Adversarial examples crafted for CIFAR-10 to fool a classifier fortified with thermometer encoding at different levels of perturbation.}}
  \label{fig:therm_diff_eps}
\end{figure}

For evaluating all defenses, we trained the substitute model for 6 substitute training epochs using CleverHans library \cite{cleverhans} and set the initial dataset to be the \red{first 150 samples in the MNIST test set for RFN, SafetyNet, and Defense-GAN and first 150 samples in the CIFAR-10 test set for thermometer encoding}. We also set $\lambda=0.1$. The success rate for the adversarial examples crafted against detecting defenses (\blue{Defense-GAN} and SafetyNet) shows the ones that could fool the classifier and are not detected by the detector. For example, for \blue{Defense-GAN}, when $\epsilon=0.5$, 92 out of 100 samples could fool the classifier, but all of them were detected as adversarial examples by the detector. The substitute model we used for attacking \blue{Defense-GAN} was Model 1 described in Table \ref{tab:model1} \blue{in the Appendix}. The substitute model used for attacking SafetyNet was the same as the one we used in our attack. We evaluated RFN by setting $\epsilon=0.25$. \gr{This is} the same value we used for evaluating our attack against RFN in 3 different scenarios, which are referred to as RFN-50, RFN-70 and RFN-90. For example, for RFN-50, 47 samples could fool the classifier more than 50 times in 100 parallel runs. For evaluating this attack against thermometer \blue{encoding}, we used the same model as the one we used in our attack for the substitute model. Figure \ref{fig:therm_diff_eps} shows the generated adversarial examples against this defense at different value of $\epsilon$. The rows are for $\epsilon=\frac{8}{255}$, $\epsilon=\frac{16}{255}$,  $\epsilon=\frac{24}{255}$, $\epsilon=\frac{32}{255}$, and  $\epsilon=\frac{64}{255}$ respectively. As can be seen, it becomes hard even for human eyes to classify these images correctly after $\epsilon=\frac{32}{255}$.
\blue{Note that for RFN with the same level of distortion, the success rate of our approach is 2 times, 2.78 times and 3.4 times better than this black-box attack for RFN-50, RFN-70 and RFN-90 respectively. For thermometer encoding, with our approach the adversary could reach 100\% success rate, but when using this black-box attack, even when $\epsilon=\frac{64}{255}$ and the images are unrecognizable by human eyes, the adversary could only reach 69\% success rate. For \gr{detecting defenses,} our approach also shows superiority and its success rate is at least 10 times better than this attack. }

\begin{table}
\centering
    \begin{tabular}{ |>{\centering\arraybackslash}m{2.0cm} | >{\centering\arraybackslash}m{2.0cm}| >{\centering\arraybackslash}m{2.0cm}| >{\centering\arraybackslash}m{1.0cm}|}
    \hline
    Defense Type & Success Rate & L2 norm & eps \\ \hline
    RFN-50 & 47\% & 5.15 & 0.25 \\ \hline
    RFN-70 & 33\% & 5.12 & 0.25 \\ \hline
    RFN-90 & 20\% & 5.11 & 0.25 \\ \hline
    Thermometer & 1\% & 1.73 &  $\frac{8}{255}$  \\ \hline
    Thermometer & 7\% & 3.40 &  $\frac{16}{255}$  \\ \hline
    Thermometer & 14\% & 5.12 &  $\frac{24}{255}$  \\ \hline
    Thermometer & 32\% & 6.83 &  $\frac{32}{255}$  \\ \hline
    Thermometer & 69\% & 13.12 &  $\frac{64}{255}$  \\ \hline
	SafetyNet & 0\% & NA & 0.2 \\ \hline
    SafetyNet & 0\% & NA & 0.3 \\ \hline
    SafetyNet & 0\% & NA & 0.4 \\ \hline
    SafetyNet & 0\% & NA & 0.5 \\ \hline
    Defense-GAN & 7\% & 6.25 & 0.3 \\ \hline
    Defense-GAN & 0\% & NA & 0.5 \\ \hline

    \end{tabular}
    \caption{\red{Success rate of Jacobian-based Data Augmentation attack against different defenses. RFN, SafetyNet, and Defense-GAN were evaluated on MNIST. Thermometer encoding was evaluated on CIFAR-10.}} \label{tab:bb_results}
\end{table}

\subsection{Transferable Adversarial Examples}

Liu et al. in \cite{transferable} designed a different targeted black box attack. In contrast to our attack, \blue{they considered a different threat model} where they don't need to query the target model to get the outputs for different inputs, and they don't need to train a substitute model. Instead of training a substitute model, they leverage an  ensemble of pre-trained models to craft adversarial examples that can be transferred to another model with their targeted labels.

The authors showed that by generating adversarial examples for an ensemble of pre-trained neural networks, it is more likely to transfer them to another classifier. Formally, given k white-box models with softmax outputs being $F_1,...,F_k$, they solve the following optimization problem:\par

{\centering $ argmin_{\delta} - \log((\sum_{i=1}^k\alpha_iF_i(x^\prime)).1_{y_{target}}) + \lambda d(x,x^\prime)$\par}

\noindent where $\sum_{i=1}^k\alpha_iF_i(x^*)$ is the ensemble model, $\alpha_i$ are the ensemble weights, and $\sum_{i=1}^k\alpha_i = 1$. $d(x,x^\prime)$ is the distance function between the original image and the perturbed image which can be $l_2$ norm.
$\lambda $  is a hyper parameter which controls the amount of distortion and the success rate. Increasing $\lambda $  results in a larger distortion, while also increasing the success rate (likelihood of being able to fool the target classifier).
Solving this optimization problem basically means that the attacker wants to maximize the score of the targeted label in all of the classifiers while keeping the amount of distortion small.

\begin{figure}
  \centering
  \includegraphics[width=45mm]{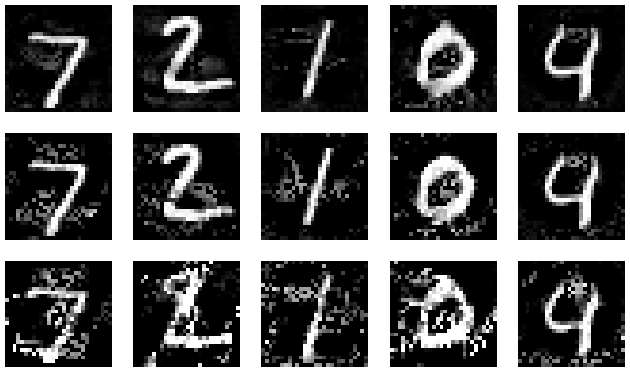}
  \setlength{\belowcaptionskip}{-15pt}
  \caption{Adversarial examples crafted with different $\lambda$ for the MNIST dataset.}
  \label{fig:trans_diff_lambda}
\end{figure}

We implemented this attack against the four defenses we considered, and the results can be found in Table \ref{tab:trans_results}. For attacking RFN, SafetyNet, and DefenseGAN, we trained four substitute models described in \gr{Tables} \ref{tab:model1}, \ref{tab:model2}, \ref{tab:model3}, and \ref{tab:model4} \red{on the MNIST} \blue{training set} for 10 epochs each and with Adam optimizer (lr=0.001). The architecture of these models can be found in the Appendix. For generating adversarial examples, as in the original paper, we used Adam with lr=0.001 to optimize the above objective and  $\alpha_i = 0.25$. For generating each adversarial example, we did 300 iterations of GD. For SafetyNet, we crafted adversarial examples by changing the $\lambda$ parameters. As you can see \gr{in Table \ref{tab:trans_results}}, when $\lambda=0.001$ the success rate is only 17\%. However, even in this case, the amount of perturbation is too high.  Samples of adversarial examples generated by this method can be found in Figure \ref{fig:trans_diff_lambda}. For the first row $\lambda$ \gr{is} 0.1, for the second one $\lambda$ \gr{is} 0.01, and for the last one $\lambda$ \gr{is} 0.001. 

For applying this attack on thermometer encoding, since our robust model was trained on \red{CIFAR-10}, we trained four other models to generate adversarial examples with them, and we trained all of them with \gr{the} CIFAR-10 training set. The models we trained for this purpose were VGG-19 \cite{vgg}, wide ResNet \cite{wideResNet}, ResNet-50 \cite{resnet} and NIN \cite{nin}. After training the VGG model reached to 93.41\% accuracy, the wide ResNet model reached to 92\% accuracy, the NIN model reached 90.29\% accuracy, and the ResNet-50 reached 94.06\% accuracy.  We trained all of these models with 4 Tesla K80 GPUs. The training time for these models were 2, 7, 1, and 6 hours respectively. This is a much longer training time than in our approach, in which we trained a smaller model, and it only took a few minutes. The robust model's accuracy was also 88.59\%. For the attack we used the same hyper parameters as the ones we used against three other defenses. Generating each adversarial example took 63 seconds on average.  Samples of crafted adversarial examples, with different $\lambda$, are shown in Figure \ref{fig:cifar_trans_diff_lambda}.
\blue{Note that for RFN, in an unbounded attack, our approach could reach 100\% success rate. Using this attack, amongst our experiments, the best case success rate was 93\% (For RFN-50 when $\lambda=0.001$). The average l2 norm of adversarial examples crafted by our approach against RFN-50 was 2.13, which is almost 3 times smaller than \gr{the} average l2 norm of crafted adversarial example using this approach. For thermometer encoding, in the best case, using this attack it could reach 39\% success rate. Using our approach, we could reach 100\% success rate, and the average l2 norm of adversarial examples found by our approach in the best case (when we used 4 substitute models) is 2.68 times smaller than \gr{the} average l2 norm of crafted adversarial examples found by this approach. For detecting defenses, our approach is also more successful and at least 5 times better than this attack.}

\begin{table}
\centering
    \begin{tabular}{ |>{\centering\arraybackslash}m{2.0cm} | >{\centering\arraybackslash}m{2.0cm}| >{\centering\arraybackslash}m{2.0cm}| >{\centering\arraybackslash}m{1.0cm}|}
    \hline
    Defense Type & Success Rate & L2 norm & $\lambda$ \\ \hline
    RFN-50 & 26\% & 1.74 & 0.1 \\ \hline
    RFN-70 & 16\% & 1.74 & 0.1 \\ \hline
    RFN-90 & 1\% & 1.59 & 0.1 \\ \hline
    RFN-50 & 84\% & 3.52 & 0.01 \\ \hline
    RFN-70 & 54\% & 3.44 & 0.01 \\ \hline
    RFN-90 & 30\% & 3.27 & 0.01 \\ \hline
    RFN-50 & 93\% & 6.31 & 0.001 \\ \hline
    RFN-70 & 88\% & 6.23 & 0.001 \\ \hline
    RFN-90 & 81\% & 6.15 & 0.001 \\ \hline
    
    Thermometer & 4\% & 2.01 &  0.1 \\ \hline
    Thermometer & 25\% & 5.11 &  0.01  \\ \hline
    Thermometer & 39\% & 7.47 &  0.001  \\ \hline

	SafetyNet & 0\% & NA & 0.1 \\ \hline
    SafetyNet & 2\% & 3.83 & 0.01 \\ \hline
    SafetyNet & 17\% & 5.58 & 0.001 \\ \hline
    SafetyNet & 17\% & 6.26 & 0.0001 \\ \hline
    Defense-GAN & 1\% & 1.62 & 0.1 \\ \hline
    Defense-GAN & 10\% & 2.78 & 0.01 \\ \hline
    Defense-GAN & 18\% & 6.33 & 0.001 \\ \hline

    \end{tabular}
      \setlength{\abovecaptionskip}{-10pt}
    \caption{\red{The success rate and average l2 norm of crafted adversarial examples by Liu et al. work against different defenses. RFN, SafetyNet, and Defense-GAN were evaluated on MNIST. Thermometer encoding was evaluated on CIFAR-10.}} \label{tab:trans_results}
\end{table}

\vspace{-0.1cm}
\begin{figure}
  \centering
  \includegraphics[width=50mm]{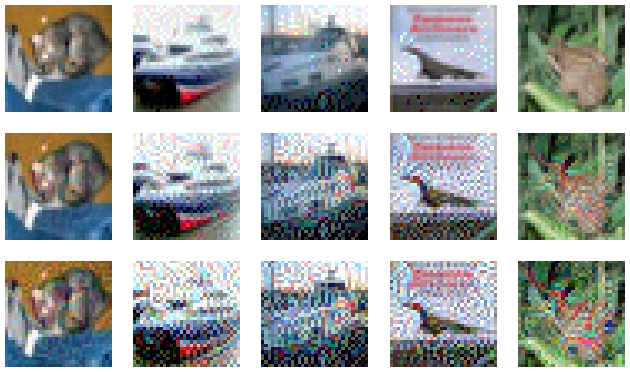}
    \setlength{\abovecaptionskip}{+10pt}
  \setlength{\belowcaptionskip}{-15pt}
  \caption{Adversarial examples crafted with different $\lambda$ for the CIFAR-10 dataset.}
  \label{fig:cifar_trans_diff_lambda}
\end{figure}

\section{Conclusion}

In this paper we described a way to craft adversarial examples against deep neural network models that leverage mechanisms to protect themselves against adversarial examples. We evaluated our approach against fortifying and detecting defenses. We showed that an adversary can craft adversarial examples without any knowledge about the type of defense used, defense parameters, model parameters, or training data.  The adversary only needs to query the robust model and train one or more substitute models. We also evaluated \blue{two black-box} attacks against the aforementioned defenses, but they performed poorly in comparison to our presented attack.  We suggest that other researchers use our approach \blue{for benchmarking in cases} where a defense prevents the attacker from calculating useful gradients from the target model.  

\blue{
\section*{Acknowledgment}
This research was supported in part by the National Science Foundation under grants 1406192 (SaTC), 1652698 (CAREER), and 1700527 (SDI-CSCS).
It also utilized the RMACC Summit supercomputer, which is supported by the National Science Foundation (awards ACI-1532235 and ACI-1532236), the University of Colorado Boulder, and Colorado State University. 
We would also like to thank the reviewers, and Nicolas Papernot in particular, for their valuable feedback and guidance.
}

\bibliographystyle{ACM-Reference-Format}
\bibliography{ccs-sample}
\section*{Appendix}
\blue{The} model architectures and parameters we used throughout the paper.
\begin{table}[H]
\centering
    \begin{tabular}{ |>{\centering\arraybackslash}m{2.0cm} | >{\centering\arraybackslash}m{3.0cm}|}
    \hline
    DNN Structure & 784-784-784-784-10 \\ \hline
    Activation & Relu \\ \hline
    Optimizer & SGD \\ \hline
    Learning Rate & 0.1 \\ \hline
    Dropout Rate & 0.25 \\ \hline
    Batch Size & 100 \\ \hline
    Epoch & 25 \\ \hline
    \end{tabular}
    \caption{The model architecture and hyper parameters used for training the model for RFN evaluation.} \label{tab:rfn_params}
\end{table}

\begin{table}[H]
\centering
    \begin{tabular}{ |>{\centering\arraybackslash}m{4.0cm} | >{\centering\arraybackslash}m{2.0cm}|}
    \hline
    Layer Type & Parameters \\ \hline
    Convolution + ReLU & $64,3 \times 3 ,1$ \\ \hline
    Convolution + ReLU & $64,3 \times 3 ,1$ \\ \hline
    Convolution + ReLU & $64,3 \times 3 ,1$ \\ \hline
    Max Pooling & $2 \times 2,2$ \\ \hline
    Convolution + ReLU & $64,3 \times 3 ,1$ \\ \hline
    Fully Connected + ReLU & $2048$ \\ \hline
    Fully Connected  & $10$ \\ \hline
    Softmax & - \\ \hline
    \end{tabular}
    \caption{The substitute model architecture and hyper parameters used for attacking RFN.} \label{tab:sub_rfn_params}
\end{table}
\vspace*{-1cm}
\begin{table}[H] \centering 
\begin{tabular}{ |>{\centering\arraybackslash}m{4.0cm} | >{\centering\arraybackslash}m{2.0cm}|} \hline
Layer Type & Parameters \\ \hline Convolution + ReLU & $64,3 \times 3 ,1$ \\ \hline 
Convolution + ReLU & $64,3 \times 3 ,1$ \\ \hline Max Pooling & $2 \times 2,2$ \\ \hline 
Convolution + ReLU & $128,3 \times 3 ,1$ \\ \hline Convolution + ReLU & $64,3 \times 3 ,1$ \\ \hline 
Max Pooling & $2 \times 2,2$ \\ \hline Convolution + ReLU & $64,3 \times 3 ,1$ \\ \hline Fully Connected + ReLU & $4096$ \\ \hline 
Fully Connected + ReLU & $1024$ \\ \hline Fully Connected & $10$ \\ \hline Softmax & - \\ \hline 
\end{tabular} \caption{The substitute model architecture and hyper parameters used for attacking Thermometer Encoding} \label{tab:therm_sub} 
\end{table}

\vspace*{-1cm}

\begin{table}[H]
\centering
    \begin{tabular}{  |>{\centering\arraybackslash}m{1.0cm} |>{\centering\arraybackslash}m{4.0cm} | >{\centering\arraybackslash}m{2.0cm}|}
    \hline
    Layer\# & Layer Type & Parameters \\ \hline
    1 & Convolution + ReLU & $64,3 \times 3,1$ \\ \hline
    2 & Max Pooling & $2 \times 2,2$ \\ \hline
    3 & Convolution + ReLU & $64,3 \times 3,1$ \\ \hline
    4 & Fully Connected + ReLU & $2048$ \\ \hline
    5 & Fully Connected  & $10$ \\ \hline
    6 & Softmax & - \\ \hline

    \end{tabular}
    \caption{The model architecture used for evaluating SafetyNet.} \label{tab:safeteynet_params}
\end{table}
\vspace*{-1cm}

\begin{table}[H]
\centering
    \begin{tabular}{  |>{\centering\arraybackslash}m{1.0cm} |>{\centering\arraybackslash}m{4.0cm} | >{\centering\arraybackslash}m{2.0cm}|}
    \hline
    Layer\# & Layer Type & Parameters \\ \hline
    1 & Convolution + ReLU & $64,3 \times 3 ,1$ \\ \hline
    2 & Convolution + ReLU & $64,3 \times 3 ,1$ \\ \hline
    3 & Convolution + ReLU & $64,3 \times 3 ,1$ \\ \hline
    4 & Max Pooling & $2 \times 2,2$ \\ \hline
    5 & Convolution + ReLU & $64,3 \times 3 ,1$ \\ \hline
    6 & Fully Connected + ReLU & $1024$ \\ \hline
    7 & Fully Connected + ReLU & $512$ \\ \hline
    8 & Fully Connected + ReLU & $512$ \\ \hline
    9 & Fully Connected  & $2$ \\ \hline
    10 & Softmax & - \\ \hline

    \end{tabular}
    \caption{The substitute detector architecture used for attacking SafetyNet.} \label{tab:safeteynet_sub_detector}
\end{table}

\vspace*{-1cm}

\begin{table}[H]
\centering
    \begin{tabular}{ |>{\centering\arraybackslash}m{4.0cm} | >{\centering\arraybackslash}m{2.0cm}|}
    \hline
     Layer Type & Parameters \\ \hline
     Convolution + ReLU & $64,5 \times 5,1$ \\ \hline
     Convolution + ReLU & $64,5 \times 5,2$ \\ \hline
     Dropout & $0.25$ \\ \hline
     Fully Connected + ReLU & $128$ \\ \hline
     Dropout & $0.5$ \\ \hline
     Fully Connected  & $10$ \\ \hline
     Softmax & - \\ \hline

    \end{tabular}
    \caption{The model architecture used for evaluating Defense-GAN.} \label{tab:dgan_model}
\end{table}
\vspace*{-1cm}

\begin{table}[H]
\centering
    \begin{tabular}{ |>{\centering\arraybackslash}m{4.0cm} | >{\centering\arraybackslash}m{2.0cm}|}
    \hline
     Layer Type & Parameters \\ \hline
     Dropout & $0.2$ \\ \hline
     Convolution + ReLU & $64,8 \times 8,2$ \\ \hline
     Convolution + ReLU & $128,6 \times 6,2$ \\ \hline
     Convolution + ReLU & $128,5 \times 5,1$ \\ \hline
     Dropout & $0.5$ \\ \hline
     Fully Connected  & $10$ \\ \hline
     Softmax & - \\ \hline

    \end{tabular}
    \caption{The Model 1 architecture.} \label{tab:model1}
\end{table}
\vspace*{-1cm}
\begin{table}[H]
\centering
    \begin{tabular}{ |>{\centering\arraybackslash}m{4.0cm} | >{\centering\arraybackslash}m{2.0cm}|}
    \hline
     Layer Type & Parameters \\ \hline
     Convolution + ReLU & $128,3 \times 3,1$ \\ \hline
     Convolution + ReLU & $64,3 \times 3,2$ \\ \hline
     Convolution + ReLU & $128,5 \times 5,1$ \\ \hline
     Dropout & $0.25$ \\ \hline
     Fully Connected + ReLU  & $128$ \\ \hline
     Dropout & $0.5$ \\ \hline
     Fully Connected  & $10$ \\ \hline
     Softmax & - \\ \hline

    \end{tabular}
    \caption{The Model 2 architecture.} \label{tab:model2}
\end{table}
\vspace*{-1cm}
\begin{table}[H]
\centering
    \begin{tabular}{ |>{\centering\arraybackslash}m{4.0cm} | >{\centering\arraybackslash}m{2.0cm}|}
    \hline
     Layer Type & Parameters \\ \hline
     Fully Connected + ReLU & $200$ \\ \hline
     Dropout & $0.5$ \\ \hline
     Fully Connected + ReLU   & $200$ \\ \hline
     Dropout & $0.5$ \\ \hline
     Fully Connected  & $10$ \\ \hline
     Softmax & - \\ \hline

    \end{tabular}
    \caption{The Model 3 architecture.} \label{tab:model3}
\end{table}
\vspace*{-1cm}
\begin{table}[H]
\centering
    \begin{tabular}{ |>{\centering\arraybackslash}m{4.0cm} | >{\centering\arraybackslash}m{2.0cm}|}
    \hline
     Layer Type & Parameters \\ \hline
     Fully Connected  + ReLU  & $200$ \\ \hline
     Fully Connected + ReLU   & $200$ \\ \hline
     Fully Connected   & $10$ \\ \hline
     Softmax & - \\ \hline

    \end{tabular}
    \caption{The Model 4 architecture.} \label{tab:model4}
\end{table}
\end{document}